\documentclass[letterpaper]{article} 
\usepackage{aaai24}  
\usepackage{times}  
\usepackage{helvet}  
\usepackage{courier}  
\usepackage[hyphens]{url}  
\usepackage{graphicx} 
\usepackage{natbib}  
\usepackage{caption} 
\usepackage{newfloat}
\usepackage{listings}
\DeclareCaptionStyle{ruled}{labelfont=normalfont,labelsep=colon,strut=off} 
\usepackage{amsmath}
\usepackage{amssymb}
\usepackage{booktabs}
\usepackage{multicol}
\usepackage{multirow}
\usepackage{algorithm2e}
\usepackage{xcolor}
\usepackage{subcaption}

\DeclareMathOperator*{\argmin}{arg\,min}

\title{Summarizing Stream Data for Memory-Constrained Online Continual Learning}
\author{
    Jianyang Gu\textsuperscript{\rm 1,2}, Kai Wang\textsuperscript{\rm 2}, Wei Jiang\textsuperscript{\rm 1}\thanks{Corresponding Author}, Yang You\textsuperscript{\rm 2}\\
}
\affiliations{
    \textsuperscript{\rm 1}Zhejiang University\\
    \textsuperscript{\rm 2}National University of Singapore\\

    \{gu\_jianyang, jiangwei\_zju\}@zju.edu.cn, \{kai.wang, youy\}@comp.nus.edu.sg
}

\usepackage{bibentry}

\begin{document}

\maketitle

\begin{abstract}
Replay-based methods have proved their effectiveness on online continual learning by rehearsing past samples from an auxiliary memory. With many efforts made on improving training schemes based on the memory, however, the information carried by each sample in the memory remains under-investigated. Under circumstances with restricted storage space, the informativeness of the memory becomes critical for effective replay. Although some works design specific strategies to select representative samples, by only employing a small number of original images, the storage space is still not well utilized. To this end, we propose to \textbf{S}ummarize the knowledge from the \textbf{S}tream \textbf{D}ata (SSD) into more informative samples by distilling the training characteristics of real images. Through maintaining the consistency of training gradients and relationship to the past tasks, the summarized samples are more representative for the stream data compared to the original images. Extensive experiments are conducted on multiple online continual learning benchmarks to support that the proposed SSD method significantly enhances the replay effects. We demonstrate that with limited extra computational overhead, SSD provides more than 3\% accuracy boost for sequential CIFAR-100 under extremely restricted memory buffer. Code in https://github.com/vimar-gu/SSD. 
\end{abstract}

\section{Introduction}

Continual learning (CL) aims to incrementally accumulate knowledge from a sequence of tasks~\cite{gu2022not,li2017learning,mccloskey1989catastrophic,ring1998child,thrun1998lifelong}.
Due to the explosive data collection by modern vision systems, storing all the images of past tasks for full fine-tuning is impractical for both computation and storage. 
Such restriction results in \textit{catastrophic forgetting}, \textit{i.e.}, adapting the model to new tasks causes knowledge forgetting and performance degradation of previous ones~\cite{mccloskey1989catastrophic}. 
In this paper, we focus on the more challenging online class-incremental CL setting, where each task consists of unique classes, with no task identity~\cite{chaudhry2018efficient,chaudhry2019tiny}. 
Inside the task, the data is passed in a non-stationary stream, which can only be accessed once. 

\begin{figure}[t]
    \centering
    \includegraphics[width=0.88\columnwidth]{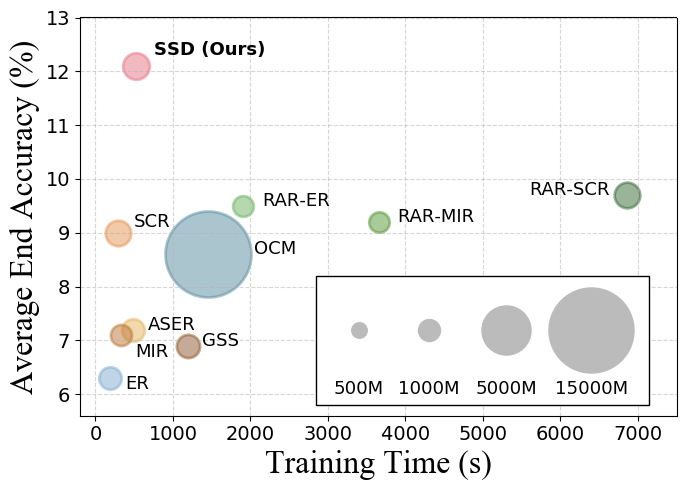}
    \caption{Under the restricted memory size of 100, the information contained in the memory is rather limited. By integrating the information from stream data into summarized samples, our proposed SSD largely enhances the replay effects. Experiments are conducted on the sequential CIFAR-100 benchmark (10 tasks). The scatter point size represents the GPU memory consumption at the training process. }
    \label{fig:intro}
\end{figure}

Recent literature addresses the forgetting problem from the perspectives of dynamic model architectures~\cite{aljundi2017expert,fernando2017pathnet}, parameter updating regularization~\cite{kirkpatrick2017overcoming,li2017learning}, pseudo sample generation~\cite{shin2017continual} and memory-assisted replay~\cite{lopez2017gradient,rebuffi2017icarl}. 
Among them, replay-based methods maintain a small auxiliary memory containing images from past tasks.
By mixing them into the current training process~\cite{chaudhry2019tiny}, it provides direct accessibility to the past knowledge. 
Such a scheme brings stable performance improvements on relatively small computational overhead. 
Many efforts have been made on improving the training schemes to better leverage the memory~\cite{mai2021supervised,zhangsimple}. 
However, the informativeness of samples in the memory remains under-investigated. 

Under circumstances with restricted storage space, the information carried by each sample in the memory becomes critical to the replay effects. 
There are some works devoted to selecting proper data from the stream for the memory~\cite{aljundi2019gradient,shim2021online}.
But by only employing a small amount of original images, the storage space is still underutilized. 
Efficiently condensing more information into the memory poses a challenging yet promising direction for further enhancing online continual learning. 

In this work, we propose a novel Summarizing Stream Data (SSD) method to integrate the information from original images into a small amount of more informative samples.
Specifically, initialized with first images in each task, the summarized samples are updated by sequentially distilling the knowledge from real images in the data stream. 
Firstly, the training gradients of real and summarized samples on the same network are matched.
Thereby training the summarized samples provides similar parameter update results to the original images. 
Secondly, we employ the samples of previous tasks in the memory to help fit the overall distribution and provide more proper gradient supervision. 
Besides, the consistency on the relationship to the previous samples between real and summarized samples also serves as a constraint for establishing better distribution in the memory. 
By maintaining the consistency on multiple aspects, the summarized samples are more representative for the task data compared with original images of the same amount. 

The proposed SSD method significantly enhances the replay effects by constructing the auxiliary memory with more informative samples. 
As shown in Fig.~\ref{fig:intro}, under circumstances with restricted memory sizes, the information lack largely influences the replay effects when only original data is employed. 
With limited extra computational overhead, the proposed SSD method boosts the average end accuracy by more than 2\%. 
We also show that SSD universally improves the performance for multiple popular CL baselines. 


\begin{figure}[t]
    \centering
    \includegraphics[width=0.9\columnwidth]{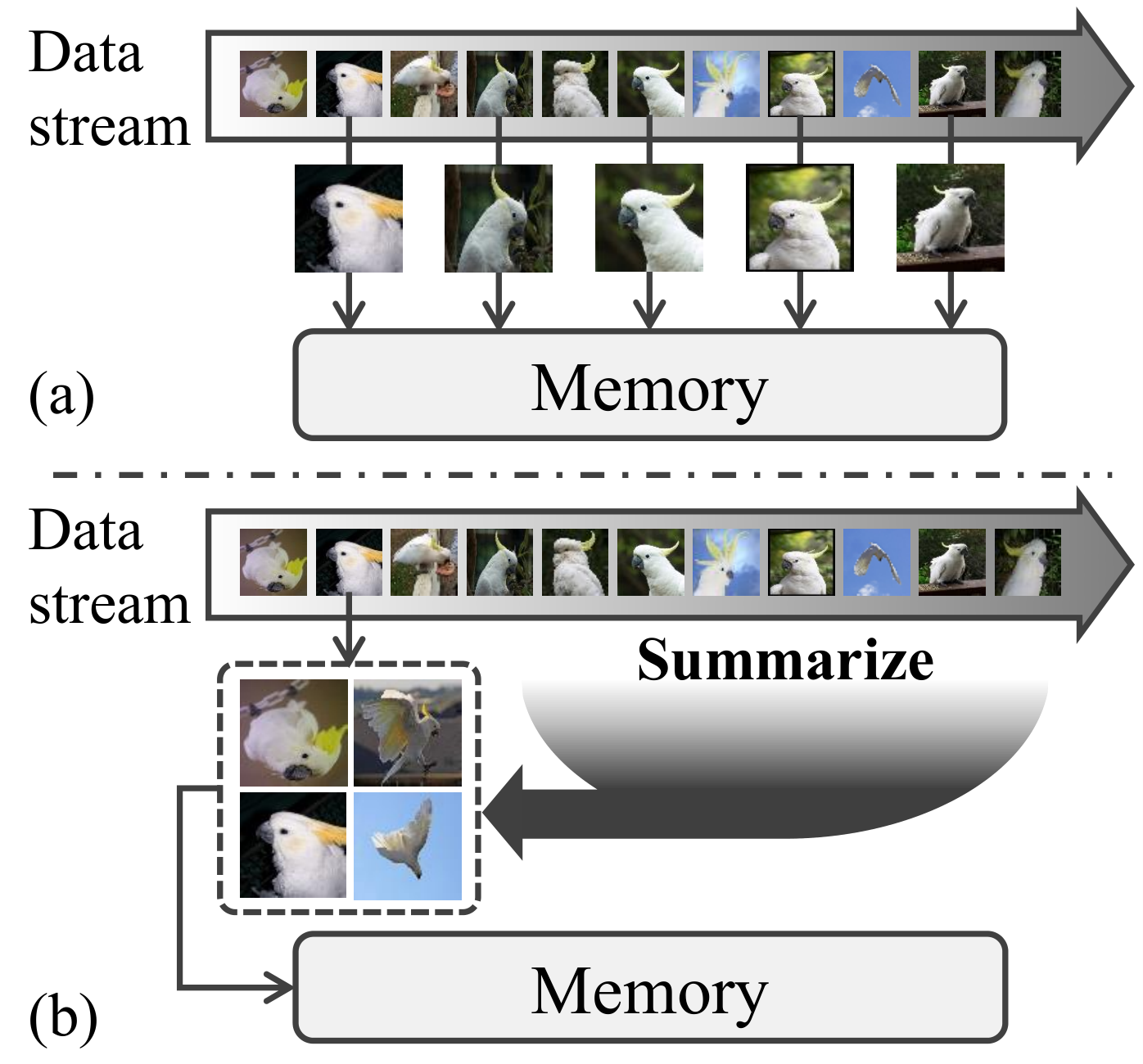}
    \caption{Concept Comparison: (a) Previous CL methods select original images from the stream data to form the auxiliary memory. (b) We propose to summarize stream data into informative samples to enhance the replay effects. }
    \label{fig:framework}
\end{figure}

\section{Related Works}
\paragraph{Continual Learning} requires a model to consecutively learn new tasks without forgetting the learned knowledge for previous tasks~\cite{mccloskey1989catastrophic,ring1998child,thrun1998lifelong}. 
Due to the restricted access to the past data, the \textit{catastrophic forgetting} issue becomes the major obstruction from practical appliance. 
Recent literature mainly addresses the problem by designing dynamic model architectures~\cite{aljundi2017expert,fernando2017pathnet,kang2022forget}, regularizing the parameter updating~\cite{kirkpatrick2017overcoming,lee2017overcoming,li2017learning}, generating pseudo samples~\cite{shin2017continual,smith2021always,xiang2019incremental}, or employing auxiliary memories for storing previous information~\cite{lopez2017gradient,rebuffi2017icarl,tiwari2022gcr}. 
Based on the differences of data increment form, CL is divided into task-incremental and class-incremental settings. 
The \textit{task-incremental} setting provides the task identity, according to which different networks can be applied. 
For the \textit{class-incremental} setting, the task identity is not provided, and different tasks share no overlapping classes. 
In this paper, we focus on the challenging class-incremental setting. 

\paragraph{Online Continual Learning} focuses on a more practical setting where the non-stationary data stream is passed only once for training~\cite{chaudhry2018efficient,phamcontinual,suninformation,prabhu2023online,harun2023siesta,ghunaim2023real}. 
Replay-based methods achieve noticeable results in online CL tasks~\cite{aljundi2019online,chaudhry2019tiny}. 
Chaudhry \textit{et al.} employ an auxiliary memory to store previous images, and re-introduce the information into the current model training~\cite{chaudhry2019tiny}. 
Aljundi \textit{et al.} select the samples that are most interfered by new incoming samples for replay~\cite{aljundi2019online}. 
DER and SCR incorporate prediction logits matching and contrastive learning for retaining the previous knowledge~\cite{buzzega2020dark,mai2021supervised}. 
Most recently, some materials further improve the performance from the perspectives of better training schemes~\cite{gu2022not,guo2022online,zhangsimple}. 

\begin{figure*}[t]
    \centering
    \includegraphics[width=0.93\textwidth]{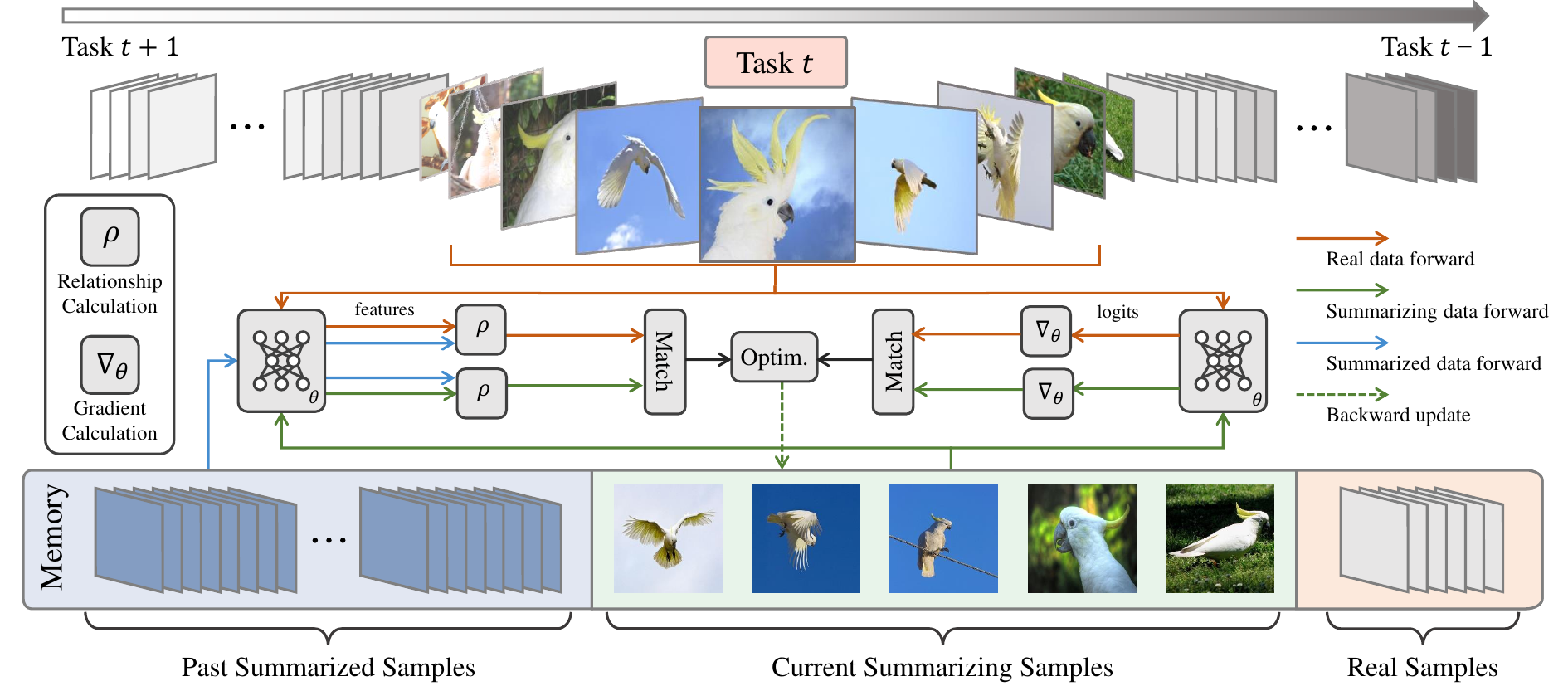}
    \caption{The pipeline of our proposed Summarizing Stream Data (SSD) method. 
    The memory is composed of the summarized samples of previous tasks (Past Summarized Samples), the samples of the current task that are being summarized (Current Summarizing Samples) and real samples. 
    For the current summarizing samples, both the training gradients and the relationships to the past knowledge are constrained to be consistent with real images. }
    \label{fig:pipeline}
\end{figure*}

With many efforts made on improving the replay training techniques, the information carried by the images inside the memory is often overlooked. 
GSS~\cite{aljundi2019gradient} selects samples with variant gradient direction to form the memory. 
Shim \textit{et al.} utilize the Shapley Value as the sample selection criterion~\cite{shim2021online}. 
In this work, we mainly focus on improving the informativeness of the memory, and propose to summarize the knowledge of the stream data into informative samples for more effective replay.  

\paragraph{Dataset Distillation} aims to integrate the information from large-scale datasets into much smaller ones for reducing the storage and calculation burdens. 
Dataset distillation (DD) methods are roughly divided into two categories: meta learning methods~\cite{nguyen2020dataset,nguyen2021dataset,looefficient,wang2018dataset}, matching-based methods~\cite{kim2022dataset,wang2022cafe,zhao2021dataset,zhao2023dataset,liu2023dream,lu2023can,du2023minimizing,liu2023slimmable} and generative methods~\cite{cazenavette2022dataset,wang2023dim,gu2023efficient}. 
By distilling the information from the whole dataset into several synthetic images, the informativeness of single sample is largely improved. 
Sangermano \textit{et al.} employs DD on CL to linearly combine the original images into a small set~\cite{sangermano2022sample}. However, the method is only eligible for datasets with simple image structures and similar backgrounds. 
Inspired by dataset distillation methods and the properties of online CL, we propose to summarize stream data into more informative samples.

\section{Method}

\subsection{Preliminaries}
In this paper, we focus on the online class-incremental continual learning (CL) setting. 
Consider a sequence of tasks $\mathcal{T}=\{\mathcal{T}_1,\mathcal{T}_2,...,\mathcal{T}_N\}$, each of which consists of unique data and classes $\mathcal{T}_t=\{(x_{ti},y_{ti})\}_{i=1}^{I_t}$, where $N$ is the total task number and $I_t$ is the data number of task $\mathcal{T}_t$. 
The class-incremental setting indicates that $\{x_{ti}\}\cap\{x_{tj}\}=\emptyset$ and $\{y_{ti}\}\cap\{y_{tj}\}=\emptyset$ for $\forall j\ne i$. 
For the offline CL setting, each task $\mathcal{T}_t$ can be accessed for multiple epochs to reach convergence. 
Yet for the online setting, each data inside the task $(x_{ti},y_{ti})$ is passed only once in a non-stationary stream, with no task identity provided. 
The target of continual learning is to maximize the overall accuracy on all the tasks. 

Replay-based methods maintain a small auxiliary memory $\mathcal{M}=\{(x_m,y_m)\}_{m=1}^K$ to store the images from past tasks, where $K$ is the memory size. 
As shown in Fig.~\ref{fig:framework} (a), most of previous methods select original samples in the stream data to form the memory. 
During each iteration of current task, a batch of samples $\mathcal{B}_m=\{(x_{mi},y_{mi})\}_{i=1}^{B_m}$ is sampled from the memory randomly or under certain strategies for joint training with current data $\mathcal{B}_t=\{(x_{ti},y_{ti})\}_{i=1}^{B_t}$, where $B_m\le K$ and $B_t$ are the mini-batch size of the replay data and the current task, respectively. 
The joint training process is formulated as follows,
\begin{equation}
\label{eq:train}
    \phi^{*}=\argmin_\phi\mathcal{L}_t(\phi;\mathcal{B}_t)+\lambda\mathcal{L}_m(\phi;\mathcal{B}_m),
\end{equation}
where $\phi$ is the parameters of the training model, $\mathcal{L}_t(\cdot;\cdot)$ is the standard training objective for the task, $\mathcal{L}_m(\cdot;\cdot)$ is the objective for the replay, and $\lambda$ is a coefficient hyper-parameter for the replay objective. 
By simultaneously viewing the data from the current and past tasks, the model can obtain new knowledge while reduce the catastrophic forgetting for the past. 
However, when the storage space is restricted, the memory informativeness becomes critical for effective replay. 
And we argue that by only employing the original images, the storage space is still underutilized. 
Therefore, we propose to Summarize Stream Data (SSD) so that the limited memory contains richer information regarding to the whole data stream. 

\subsection{Summarizing Stream Data}
Instead of selecting original samples from the stream data, as shown in Fig.~\ref{fig:framework} (b), we propose to integrate the information from the whole stream data into several informative images. 
By maintaining summarized samples that better represent the corresponding data distribution, the more informative auxiliary memory helps improve the replay effects. 
Even under restricted memory size, the proposed SSD method provides stable performance improvements on the average accuracy over the past tasks. 
The detailed pipeline of the proposed SSD method is presented in Fig.~\ref{fig:pipeline}. 

\paragraph{Dynamic memory.}
We first design a dynamic memory to maintain both summarized and original samples. 
When the data stream brings in new classes, a small specified number of memory will be initialized with the first images of the corresponding classes. 
In previous replay-based methods, images stored in any memory position can be replaced with new ones. 
As we intend to summarize as much information from the data stream as possible, these initialized positions are fixed to only be updated with summarizing information instead of direct replacement. 
For the other memory positions, standard reservior update is conducted as in~\cite{chaudhry2019tiny}. 
Thereby the memory space is not wasted when the experienced task number is small. 
Under circumstances with restricted space, it is important to keep the example number per class balanced, as in~\cite{rebuffi2017icarl,yoon2022online}. 
Thus the initialization position number for each class $k$ is pre-defined to be the memory size $K$ divided by the total class number. 
For convenience, we define the summarized images and original images in the memory as $\mathcal{M}_S$ and $\mathcal{M}_O$ in the follows. 

\paragraph{Data summarizing.}
The main purpose of the proposed SSD method is to increase the richness of the information carried by the stored samples and to minimize the distance between the memory and the original data distribution to obtain the optimized memory $\mathcal{M}^*$:
\begin{equation}
    \mathcal{M}^*=\argmin_{\mathcal{M}}\mathbf{D}(\mathcal{M},\mathcal{T}),
\end{equation}
where $\mathbf{D(\cdot,\cdot)}$ represents the distance metric. 
We expect the informative samples to have similar training responses to the whole original image set. 
Inspired by dataset distillation methods, given the stream data, we implement the summarizing process by sequentially distilling the knowledge from mini-batches of real data into the summarized samples of the corresponding classes~\cite{zhao2021dataset}. 

We define the summarized samples of class $c$ as $\mathcal{M}_c$, whose size is pre-defined as $k$, and the data of the same class in the current mini-batch as $\mathcal{B}_c$. 
The optimization target of the summarizing process is to minimize the distance between $\mathcal{M}_c$ and $\mathcal{B}_c$. 
DC~\cite{zhao2021dataset} proposes a simple yet effective metric that is the distance of the training gradients on the same network between the summarized and original images. 
By matching the model updating metrics throughout the stream data, the summarized samples are updated to approach the training effects of the whole real dataset. 
The gradient matching objective is formulated as follows,
\begin{equation}
    \mathcal{L}_g=\mathbf{D}\left(\nabla_\theta \mathcal{L}_t^{'}(\theta;\mathcal{M}_c),\nabla_\theta \mathcal{L}_t^{'}(\theta;\mathcal{B}_c)\right),
\end{equation}
where $\mathcal{L}_t^{'}(\cdot;\cdot)$ here is a training objective that differs from the standard training objective in Eq.~\ref{eq:train}, and $\theta$ is the parameters of the model employed for summarizing. 

In order to save extra computational overhead, the summarizing is conducted every $\tau$ iterations. 
We provide the ablation study on the summarizing interval in Fig.~\ref{fig:interval}. 

\begin{table}[t]
\begin{center}
\small
\setlength{\tabcolsep}{8pt}
\begin{tabular}{ccccc}
\toprule
\multirow{2}{*}{Slices} & \multicolumn{4}{c}{CIFAR-100} \\
 & \textbf{100$\times$1} & 50$\times$2 & 20$\times$5 & 10$\times$10 \\
\midrule
Acc & \textbf{51.6$_{\pm0.4}$} & 51.2$_{\pm0.4}$ & 50.7$_{\pm0.6}$ & 48.9$_{\pm0.6}$ \\
\bottomrule
\end{tabular}
\caption{Evaluation results of images separately distilled from different slices on CIFAR-100. Experiments conducted under the setting of 50 images per class. }
\label{tab:cifar}
\end{center}
\end{table}


\subsection{Past-assisted Summarizing}
In addition to the training gradients, we propose to take better advantage of the past accumulated knowledge to help the information summarizing of the current task. 

\paragraph{Summarizing model training.}
In the standard dataset distillation process, the gradient matching is accompanied with model updating in an alternative manner.
As complete a training process as possible is simulated in order that the gradients of different training stages can be matched. 
The parameters $\theta$ is first randomly initialized. 
At each training iteration where a new batch of stream data $\mathcal{B}_t$ is presented, $\theta$ is updated as follows,
\begin{equation}
    \theta\leftarrow\theta-\eta\nabla_\theta\mathcal{L}^{'}_t(\theta;\mathcal{B}_t),
\end{equation}
where $\eta$ is the learning rate for the summarizing model. 
Only the real images participate in the update of the summarizing model, which prevents the knowledge leakage of the summarized samples. 

For the continual learning, however, the class number is not fixed throughout the training process. 
Thus, the model has to be re-initialized when new classes come. 
As we conduct the experiments under the class-incremental setting, past classes will be no more presented in the stream data. 
If the summarizing model only constructs the decision boundaries for the current classes, the provided training gradients will lose insights on the past knowledge. 
We provide an empirical evidence here of conducting dataset distillation on different slices of CIFAR-100 in Tab.~\ref{tab:cifar}. 
It suggests that the model trained with more classes at the same time provides more proper gradient supervision. 

Therefore, we propose to employ the real images in the memory $\mathcal{M}_O$ to assist the summarizing model update:
\begin{equation}
\label{eq:theta}
    \theta\leftarrow\theta-\eta\nabla_\theta\left(\mathcal{L}^{'}_t(\theta;\mathcal{B}_t)+\mathcal{L}^{'}_t(\theta;\mathcal{M}_O)\right).
\end{equation}

\paragraph{Relationship matching.}
Since the model has obtained rough sketch on the past data, we further introduce a relationship matching to enhance the summarizing effects. 
Setting the extracted features of previous summarized samples as anchors, we explicitly constrain the mean features of summarized samples and real images to have consistent relationship to the anchors.
The objective for relationship matching is formulated as:
\begin{equation}
    \mathcal{L}_r=\mathbf{D}\left(\rho(\mathcal{M}_c, \mathcal{M}_S\backslash\mathcal{M}_c,\theta),\rho(\mathcal{B}_c,\mathcal{M}_S\backslash\mathcal{M}_c,\theta)\right),
\end{equation}
\begin{equation}
    \rho(\mathcal{X},\mathcal{Y},\theta)=\mathbf{D}\left(\overline{\mathcal{F}(\theta;\mathcal{X})},\mathcal{F}(\theta;\mathcal{Y})\right),
\end{equation}
where $\rho$ represents the relationship calculation, $\mathcal{M}_S\backslash\mathcal{M}_c$ stands for the other summarized samples except for $\mathcal{M}_c$, $\mathcal{F}(\cdot;\cdot)$ represents the feature extraction process, and $\overline{\mathcal{F}}$ is the mean feature. 
The relationship consistency helps establish better overall distribution of summarized samples inside the memory. 
Combining the gradient matching, the complete summarizing objective $\mathcal{L}_s$ is formulated as:
\begin{equation}
\label{eq:update}
    \mathcal{L}_s=\mathcal{L}_g+\gamma\mathcal{L}_r,
\end{equation}
where $\gamma$ is the coefficient for relationship matching.

\section{Experiments}

\begin{table*}[t]
\begin{center}
\small
\setlength{\tabcolsep}{5pt}
\begin{tabular}{cccc|ccc|ccc}
\toprule
\multirow{2}{*}{Method} & \multicolumn{3}{c|}{Mini-ImageNet} & \multicolumn{3}{c|}{CIFAR-100} & \multicolumn{3}{c}{Tiny-ImageNet} \\
 & $K$=100 & $K$=500 & $K$=1000 & $K$=100 & $K$=500 & $K$=1000 & $K$=200 & $K$=1000 & $K$=2000 \\
 \midrule
 fine-tune & \multicolumn{3}{c|}{4.2$_{\pm0.3}$} & \multicolumn{3}{c|}{5.8$_{\pm0.2}$} & \multicolumn{3}{c}{2.3$_{\pm0.2}$} \\
 iid offline & \multicolumn{3}{c|}{47.7$_{\pm0.2}$} & \multicolumn{3}{c|}{50.2$_{\pm0.3}$} & \multicolumn{3}{c}{25.9$_{\pm0.4}$} \\
 EWC++~\cite{chaudhry2018riemannian} & \multicolumn{3}{c|}{4.7$_{\pm0.5}$} & \multicolumn{3}{c|}{5.2$_{\pm0.4}$} & \multicolumn{3}{c}{2.1$_{\pm0.3}$} \\
 LwF~\cite{li2017learning} & \multicolumn{3}{c|}{9.6$_{\pm0.6}$} & \multicolumn{3}{c|}{12.2$_{\pm0.6}$} & \multicolumn{3}{c}{6.0$_{\pm0.3}$} \\
\midrule
AGEM~\cite{chaudhry2018efficient} & 5.0$_{\pm0.5}$ & 4.5$_{\pm0.4}$ & 4.5$_{\pm0.6}$ & 5.6$_{\pm0.4}$ & 5.7$_{\pm0.3}$ & 5.4$_{\pm0.4}$ & 2.3$_{\pm0.4}$ & 2.2$_{\pm0.3}$ & 2.4$_{\pm0.3}$ \\

ER~\cite{chaudhry2019tiny} & 5.7$_{\pm0.6}$ & 7.1$_{\pm0.7}$ & 7.6$_{\pm0.7}$ & 6.3$_{\pm0.4}$ & 9.3$_{\pm0.6}$ & 10.1$_{\pm1.1}$ & 2.3$_{\pm0.3}$ & 2.5$_{\pm0.2}$ & 2.7$_{\pm0.2}$ \\

MIR~\cite{aljundi2019online} & 6.1$_{\pm0.5}$ & 8.8$_{\pm0.6}$ & 10.6$_{\pm1.2}$ & 7.1$_{\pm0.3}$ & 9.7$_{\pm0.4}$ & 11.3$_{\pm0.5}$ & 2.5$_{\pm0.5}$ & 2.9$_{\pm0.3}$ & 3.0$_{\pm0.3}$ \\

GSS~\cite{aljundi2019gradient} & 5.9$_{\pm0.6}$ & 7.3$_{\pm0.8}$ & 7.3$_{\pm0.7}$ & 6.9$_{\pm0.5}$ & 9.1$_{\pm0.5}$ & 11.5$_{\pm0.4}$ & 2.4$_{\pm0.4}$ & 2.6$_{\pm0.3}$ & 2.8$_{\pm0.4}$ \\

ASER$_\mu$~\cite{shim2021online} & 6.4$_{\pm0.4}$ & 12.6$_{\pm0.6}$ & 15.2$_{\pm0.6}$ & 7.2$_{\pm0.3}$ & 10.3$_{\pm0.6}$ & 13.5$_{\pm0.7}$ & 2.8$_{\pm0.4}$ & 4.2$_{\pm0.4}$ & 5.4$_{\pm0.5}$ \\

DVC~\cite{gu2022not} & 8.3$_{\pm0.5}$ & 12.6$_{\pm1.0}$ & 15.2$_{\pm0.8}$ & 10.4$_{\pm0.4}$ & 16.6$_{\pm0.6}$ & 20.2$_{\pm0.4}$ & 3.4$_{\pm0.2}$ & 5.1$_{\pm0.3}$ & 6.8$_{\pm0.4}$ \\

SCR~\cite{mai2021supervised} & 8.3$_{\pm0.4}$ & 18.0$_{\pm0.6}$ & 24.4$_{\pm0.3}$ & 9.0$_{\pm0.5}$ & 20.6$_{\pm0.4}$ & 26.6$_{\pm0.6}$ & 3.0$_{\pm0.2}$ & 5.8$_{\pm0.2}$ & 8.3$_{\pm0.3}$ \\

OCM~\cite{guo2022online} &  -  &  -  &  -  & 8.6$_{\pm0.5}$ & 18.6$_{\pm0.4}$ & 28.2$_{\pm0.6}$ & 2.5$_{\pm0.3}$ & \textbf{7.7$_{\pm0.6}$} & \textbf{9.9$_{\pm0.5}$} \\

RAR~\cite{zhangsimple} & 9.1$_{\pm0.6}$ & 19.6$_{\pm0.7}$ & \textbf{26.2$_{\pm0.6}$} & 9.7$_{\pm0.3}$ & 21.2$_{\pm0.5}$ & 28.5$_{\pm0.4}$ & 3.5$_{\pm0.4}$ & 6.7$_{\pm0.7}$ & 9.2$_{\pm0.6}$ \\

SSD (Ours) & \textbf{10.3$_{\pm0.6}$} & \textbf{19.8$_{\pm0.4}$} & 25.8$_{\pm0.3}$ & \textbf{12.1$_{\pm0.4}$} & \textbf{23.0$_{\pm0.6}$} & \textbf{28.8$_{\pm0.5}$} & \textbf{3.6$_{\pm0.2}$} & 6.4$_{\pm0.5}$ & 8.9$_{\pm0.3}$ \\
\bottomrule
\end{tabular}
\caption{Average end accuracy on Mini-ImageNet (10 tasks), CIFAR-100 (10 tasks) and Tiny-ImageNet (20 tasks) with different memory sizes $K$. The averaged results of 10 runs are reported. The upper part shows results of memory-agnostic methods, and the lower part contains results of replay-based methods. }
\label{tab:sota}
\end{center}
\end{table*}

\begin{table}[t]
\begin{center}
\small
\setlength{\tabcolsep}{9pt}
\begin{tabular}{cccccc}
\toprule
\multicolumn{3}{c}{Modules} & \multicolumn{3}{c}{CIFAR-100} \\
 D & S & P & $K$=100 & $K$=500 & $K$=1000 \\
\midrule
- & - & - & 9.0$_{\pm0.5}$ & 20.6$_{\pm0.4}$ & 26.6$_{\pm0.6}$ \\
\checkmark & - & - & 10.7$_{\pm0.5}$ & 21.3$_{\pm0.4}$ & 27.2$_{\pm0.7}$ \\
\checkmark & \checkmark & - & 11.5$_{\pm0.7}$ & 22.3$_{\pm0.6}$ & 28.2$_{\pm0.6}$ \\
\checkmark & \checkmark & \checkmark & \textbf{12.1$_{\pm0.4}$} & \textbf{23.0$_{\pm0.6}$} & \textbf{28.8$_{\pm0.5}$} \\
\bottomrule
\end{tabular}
\caption{Ablation study on the effectiveness of each component in SSD. D: dynamic memory; S: summarizing data stream; P: past assistance. $K$ refers to the memory size. }
\label{tab:component}
\end{center}
\end{table}



\subsection{Datasets and Evaluation Metric}
We evaluate our methods on three popular continual learning benchmarks. 
\textbf{Sequential CIFAR-100} splits the CIFAR-100 dataset into 10 tasks, each with 10 non-overlapping classes~\cite{krizhevsky2009learning}. 
\textbf{Sequential Mini-ImageNet} splits the Mini-ImageNet dataset into 10 tasks, and each task contains 10 classes~\cite{vinyals2016matching}. 
\textbf{Sequential Tiny-ImageNet} splits the Tiny-ImageNet dataset into 20 tasks, each of which consists of 10 independent classes~\cite{deng2009imagenet}. 
For performance comparison, we mainly report the average end accuracy with variance of all tasks when the whole training process is over and each experiment is conducted 10 times. 

\subsection{Implementation Details}
For training the sequential tasks, following~\cite{mai2021supervised}, we adopt a reduced ResNet-18~\cite{he2016deep} architecture for all the experiments. 
In order to simulate the circumstances with restricted memory space, we set the memory size to contain 1, 5 and 10 images per class. 
We also extend the memory space to more ideal settings in the ablation study. 
An SGD optimizer is adopted for parameter updating, with the learning rate set as 0.1. 
By default, SCR~\cite{mai2021supervised} is adopted as the baseline, together with its training settings including objectives $\mathcal{L}_t$ and $\mathcal{L}_m$. 

For the stream data summarizing, a 3-layer ConvNet is employed to extract the features and calculate the training gradients~\cite{zhao2021dataset,kim2022dataset}. 
An SGD optimizer with a learning rate $\eta$ of 0.01 and a momentum of 0.9 is employed to update the summarizing model as in Eq.~\ref{eq:theta}. The training objective $\mathcal{L}^{'}_t(\cdot;\cdot)$ is set as the standard cross entropy loss. 
The distance metric $\mathbf{D}$ is set as the euclidean distance. 
The summarizing interval $\tau$ is set as 6 and the similarity matching coefficient $\gamma$ is set as 1.

\subsection{Comparison with State-of-the-art Methods}
We conduct the standard online class-incremental continual learning experiments on restricted memory sizes in Tab.~\ref{tab:sota}. 
All the results are reproduced by us. 
For RAR~\cite{zhangsimple}, we reproduce its results on SCR~\cite{mai2021supervised} for fair comparison. 
The result of OCM~\cite{guo2022online} on Mini-ImageNet is not provided in the original paper, and constrained by computational overhead, we are not able to reproduce it. 

We first present the results of some memory-agnostic methods in the upper part of the table. 
``fine-tune'' stands for naive CL without addressing the catastrophic forgetting problem, and ``iid offline'' represents training all the data together for 50 epochs, acting as the upper bound. 
The results of replay-based methods are presented in the lower part. 
Under circumstances with extremely small memory size ($K$=100), the performance difference between methods is relatively small compared to larger sizes. 
It indicates that the limited information contained in the memory largely influences the replay effects. 
Among the replay-based methods, SSD is the only method achieving comparable performance with LwF~\cite{li2017learning} under a restricted memory size of 100 on CIFAR-100 and Mini-ImageNet benchmarks. 
Besides, for all the memory sizes, SSD surpasses the baseline method (SCR), proving that SSD manages to improve the informativeness of the memory. 
Compared with state-of-the-art methods OCM and RAR, SSD performs better on small memory size and comparable on larger sizes with much less computational cost, as shown in Fig.~\ref{fig:intro}. 

\begin{table}[t]
\begin{center}
\small
\setlength{\tabcolsep}{2.5pt}
\begin{tabular}{cccc}
\toprule
\multirow{2}{*}{Method} & \multicolumn{3}{c}{CIFAR-100} \\
 & $K$=100 & $K$=500 & $K$=1000 \\
\midrule
ER~\cite{chaudhry2019tiny} & 6.3 & 9.3 & 10.1 \\
+SSD & \textbf{7.5} {(+1.2)} & \textbf{10.8} {(+1.5)} & \textbf{12.9} {(+2.8)} \\
\midrule
DVC~\cite{gu2022not} & 10.4 & 16.6 & 20.2 \\
+SSD & \textbf{12.0} {(+1.6)} & \textbf{18.5} {(+1.9)} & \textbf{21.8} {(+1.6)} \\
\midrule
SCR~\cite{mai2021supervised} & 9.0 & 20.6 & 26.6 \\
+SSD & \textbf{12.1} {(+3.1)} & \textbf{23.0} {(+2.4)} & \textbf{28.8} {(+2.2)} \\
\bottomrule
\end{tabular}
\caption{The experiment results of applying SSD on multiple baseline methods. $K$ refers to the memory size. }
\label{tab:baseline}
\end{center}
\end{table}

\begin{figure*}[t]
\begin{subfigure}[t]{0.24\textwidth}
    \centering
    \includegraphics[width=\textwidth]{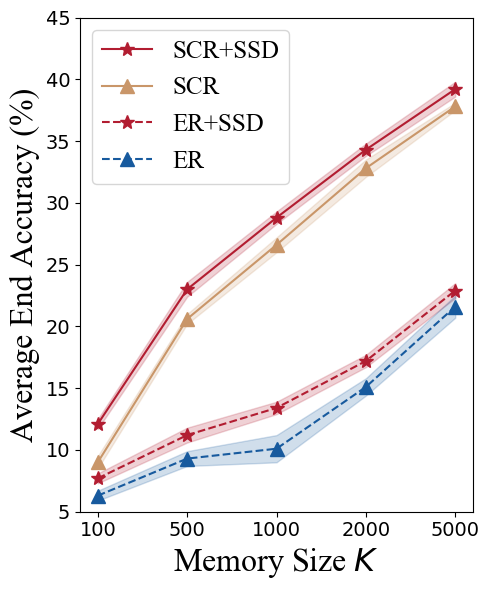}
    \caption{}
    \label{fig:memory}
\end{subfigure}
\begin{subfigure}[t]{0.24\textwidth}
    \centering
    \includegraphics[width=\textwidth]{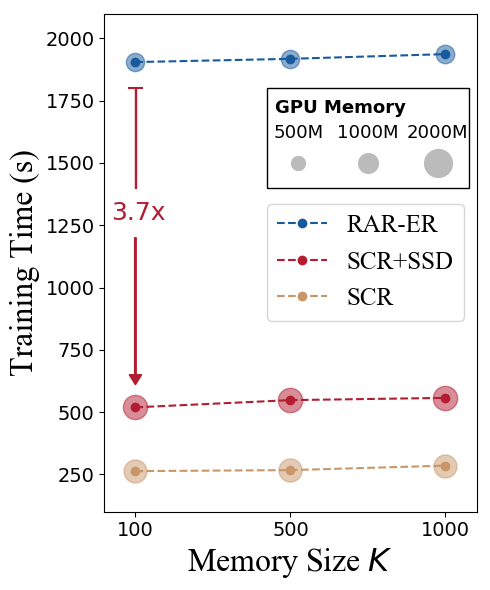}
    \caption{}
    \label{fig:comp}
\end{subfigure}
\begin{subfigure}[t]{0.24\textwidth}
    \centering
    \includegraphics[width=\textwidth]{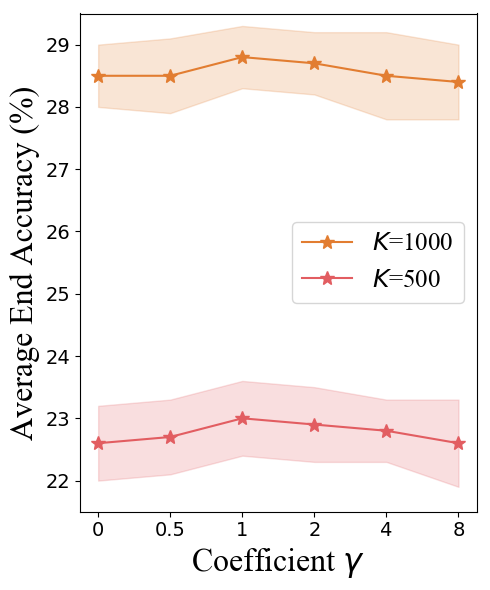}
    \caption{}
    \label{fig:coefficient}
\end{subfigure}
\begin{subfigure}[t]{0.265\textwidth}
    \centering
    \includegraphics[width=\textwidth]{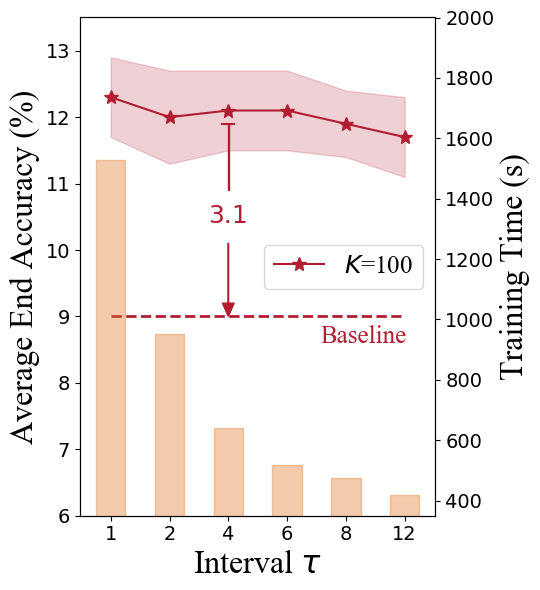}
    \caption{}
    \label{fig:interval}
\end{subfigure}
\caption{(a) The experiment results of increasing the memory size. (b) The computational cost comparison. (c) The parameter analysis on the relationship matching coefficient $\gamma$. (d) The parameter analysis on the summarizing interval $\tau$. }
\end{figure*}

\subsection{Ablation Study and Analysis}

\paragraph{Component analysis.}
We present the experiment results of adding each proposed component to the baseline method in Tab.~\ref{tab:component}. 
With SCR~\cite{mai2021supervised} set as the baseline, we first construct the dynamic memory. 
When the memory space is restricted, a class-balanced sampler becomes essential to provide necessary knowledge rehearsal for each class. 
With class-balanced knowledge, the dynamic memory achieves certain improvements on all memory sizes. 
But the margin is decayed as the memory size grows. 

Then we introduce the proposed summarizing operation with training gradient matching to integrate information into memory samples. 
The summarized samples provide steady performance gains over the dynamic memory, which proves the effectiveness of the summarizing process.
In addition to the training gradients, we also employ past summarized samples to provide better gradient supervision and global distribution guidance to the current summarizing process. 
The past-assisted summarizing further promotes the total accuracy margin to more than 2\% over the SCR baseline with limited extra computational cost. 

\begin{table}[t]
\begin{center}
\small
\setlength{\tabcolsep}{4.5pt}
\begin{tabular}{cccc}
\toprule
\multirow{2}{*}{Method} & \multicolumn{3}{c}{CIFAR-100} \\
 & $K$=100 & $K$=500 & $K$=1000 \\
\midrule
SCR~\cite{mai2021supervised} & 9.0 & 20.6 & 26.6 \\
SCR+GSS & 9.9 {(+0.9)} & 19.3 {(-1.3)} & 24.4 {(-2.2)} \\
SCR+ASER & 11.7 {(+2.7)} & 21.8 {(+1.2)} & 26.9 {(+0.3)} \\
Ours & \textbf{12.1} {(+3.1)} & \textbf{23.0} {(+2.4)} & \textbf{28.8} {(+2.2)} \\
\bottomrule
\end{tabular}
\caption{Comparisons between the proposed SSD method and other memory construction methods. }
\label{tab:memory_cons}
\end{center}
\end{table}

\paragraph{Appliance on other baselines.}
The proposed summarizing method is orthogonal to most of the previous continual learning methods. 
In order to validate the generalization capability of SSD, we apply it on multiple baselines in Tab.~\ref{tab:baseline}. 
On all the baselines SSD achieves stable performance gains by improving the informativeness of the memory. 
We also observe that with more appropriate memory replay training method, SSD generally obtains higher performance gain. 
With more rational training schemes, the more informative memory is even better utilized. 

\paragraph{Comparing memory-construction methods.}
There are also some previous works devoted to constructing more effective auxiliary memories. We compare the performance of applying them to the same SCR baseline in Tab.~\ref{tab:memory_cons}. 
GSS~\cite{aljundi2019gradient} adopts a greedy strategy to maximize the gradient variance inside the memory. 
ASER~\cite{shim2021online} selects samples with high Shapley Values to form the memory. 
The results suggest that all the methods improve the memory informativeness under extremely small memory size.
When the memory size is gradually enlarged, GSS fails to provide representative samples, and the performance margin of ASER also degrades fast. 
By contrast, SSD provides information gain across all memory sizes. 

\paragraph{Scalability to larger memory.}
In addition to the circumstances with restricted memory space, CL can also be applied on more ideal environments with sufficient storage. 
We validate the scalability of the proposed SSD method in Fig.~\ref{fig:memory}. 
As the memory space increases, SSD stably provides performance improvements on both ER and SCR baselines. 
It validates from the performance aspect that SSD is able to be applied under larger memory space. 

\begin{table}[t]
\begin{center}
\small
\setlength{\tabcolsep}{3.5pt}
\begin{tabular}{cccc}
\toprule
\multirow{2}{*}{Method} & \multicolumn{3}{c}{CIFAR-100} \\
 & $K$=100 & $K$=500 & $K$=1000 \\
\midrule
RAR~\cite{zhangsimple} & 9.7 & 21.2 & 28.5 \\
SSD & 12.1 {(+2.4)} & 23.0 {(+1.8)} & 28.8 {(+0.3)} \\
SSD-5 & 13.5 {(+3.8)} & 24.4 {(+3.2)} & 30.4 {(+1.9)} \\
SSD-10 & \textbf{13.7} {(+4.0)} & \textbf{24.7} {(+3.5)} & \textbf{31.1} {(+2.6)} \\
\bottomrule
\end{tabular}
\caption{The experiment results of extending training iterations. SSD-5 and SSD-10 indicate 5 and 10 replays inside each iteration, respectively. }
\label{tab:more_iter}
\end{center}
\end{table}

\paragraph{Computational cost analysis.}
The proposed SSD method significantly improves the informativeness of samples in the auxiliary memory. 
Compared to previous CL methods where original samples are selected from the data stream, SSD involves a distilling operation to integrate the information into memory samples. 
We analyze the required extra computational overhead in Fig.~\ref{fig:comp}. 
As the memory size increases, although there are more samples that require to be optimized, the extra computational cost still remains stable. 
Besides, SSD requires much less training time than RAR~\cite{zhangsimple} yet achieves better or comparable results. 
It validates from the computational overhead that the proposed SSD method is capable for real-world appliance. 

\paragraph{Extending training iterations.}
RAR~\cite{zhangsimple} implements the CL training process with more memory replays inside each iteration, which is helpful for retaining the past knowledge. 
We also apply the repeated replay into our SSD method in Tab.~\ref{tab:more_iter}. 
Through increasing the replay iterations, the rich information carried by summarized samples is better utilized, which further enlarges the performance gain of SSD. 
When the memory has larger space and contains more diverse information, the repeated play provides more obvious improvements. 

\begin{figure}[t]
    \centering
    \includegraphics[width=0.9\columnwidth]{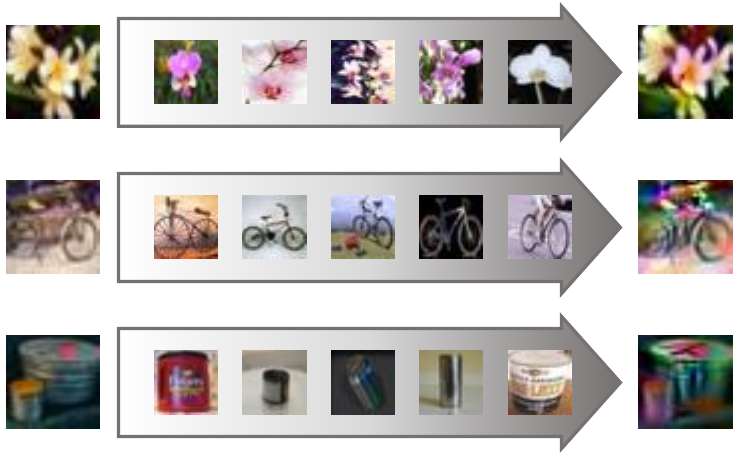}
    \caption{Visualizations of original images (left), data stream (middle) and summarized images (right) on CIFAR-100. The information of color, structure and texture are integrated into the summarized images, helping improve the replay effects. Best viewed in color. }
    \label{fig:vis}
\end{figure}

\subsection{Visualization}
We provide the comparison between the original images and the summarized ones in Fig.~\ref{fig:vis} to explicitly explain the effectiveness of the proposed summarizing method. 
Generally, more diverse colors that appear in the real distribution are introduced into the summarized images. 
For orchid (the first row) specifically, the initialized flowers are of yellow color, while most of real samples are pink. 
By modifying part of the yellow flowers into pink, the bias is corrected, and the original yellow features are not completely eliminated. 
For bicycles and cans, the structure and texture information are largely enhanced during the summarizing process, respectively. 
The summarized images contain much richer information compared with the original ones, which helps improve the replay effects. 
Note that SSD is different from the color jittering augmentation, where the color is modified in a random manner. 
More visualizations are presented in the supplementary material. 

\subsection{Parameter Analysis}

\paragraph{Relationship Matching Coefficient. }
Training gradients and relationship to the past knowledge simultaneously serve as the matching metrics for distilling the information from the data stream. 
From the results in Fig.~\ref{fig:coefficient}, we can observe that adding relationship matching generally improves the average accuracy. 
And the best results are achieved when setting the coefficient $\gamma=1$. 

\paragraph{Summarizing Interval. }
\label{sec:interval}
The interval $\tau$ influences the frequency of distilling information, and the calculation time of CL. 
As shown in Fig.~\ref{fig:interval},
SSD achieves the best performance when conducting summarizing at each iteration. 
However, it requires a large amount of calculation time. 
As the interval increases from 2 to 6, there is no obvious performance drop. 
Based on the comprehensive consideration of performance and computational overhead, we set the interval $\tau=6$. 
Under even larger summarizing intervals, SSD still achieves higher accuracy compared to the baseline.

\section{Conclusion}
In this paper, we propose a Summarizing Stream Data (SSD) method to improve the informativeness of the auxiliary memory for replay-based online continual learning methods. 
Through sequentially integrating the information from the stream data, the summarized samples are more representative for the whole data distribution compared with original images. 
With limited extra computational overhead, SSD helps significantly improve the replay effects, especially under circumstances with restricted memory space. 
The efficient and effective summarizing method inspires future works to design better means to constructing memories. 

\section{Acknowledgments}
This work was supported in part by the National Natural Science Foundation of China under Grant 62173302 and in part by the Zhejiang Provincial Natural Science Foundation of China under Grant Z24F030007. Yang You's research group is being sponsored by NUS startup grant (Presidential Young Professorship), Singapore MOE Tier-1 grant, ByteDance grant, ARCTIC grant, SMI grant, Alibaba grant, and Google grant for TPU usage.

\bibliography{aaai24}

\begin{thebibliography}{52}
\providecommand{\natexlab}[1]{#1}

\bibitem[{Aljundi et~al.(2019{\natexlab{a}})Aljundi, Belilovsky, Tuytelaars, Charlin, Caccia, Lin, and Page-Caccia}]{aljundi2019online}
Aljundi, R.; Belilovsky, E.; Tuytelaars, T.; Charlin, L.; Caccia, M.; Lin, M.; and Page-Caccia, L. 2019{\natexlab{a}}.
\newblock Online Continual Learning with Maximal Interfered Retrieval.
\newblock In \emph{NeurIPS}, 11849--11860.

\bibitem[{Aljundi, Chakravarty, and Tuytelaars(2017)}]{aljundi2017expert}
Aljundi, R.; Chakravarty, P.; and Tuytelaars, T. 2017.
\newblock Expert gate: Lifelong learning with a network of experts.
\newblock In \emph{CVPR}, 3366--3375.

\bibitem[{Aljundi et~al.(2019{\natexlab{b}})Aljundi, Lin, Goujaud, and Bengio}]{aljundi2019gradient}
Aljundi, R.; Lin, M.; Goujaud, B.; and Bengio, Y. 2019{\natexlab{b}}.
\newblock Gradient based sample selection for online continual learning.
\newblock \emph{NeurIPS}, 32.

\bibitem[{Buzzega et~al.(2020)Buzzega, Boschini, Porrello, Abati, and Calderara}]{buzzega2020dark}
Buzzega, P.; Boschini, M.; Porrello, A.; Abati, D.; and Calderara, S. 2020.
\newblock Dark experience for general continual learning: a strong, simple baseline.
\newblock \emph{NeurIPS}, 33: 15920--15930.

\bibitem[{Cazenavette et~al.(2022)Cazenavette, Wang, Torralba, Efros, and Zhu}]{cazenavette2022dataset}
Cazenavette, G.; Wang, T.; Torralba, A.; Efros, A.~A.; and Zhu, J.-Y. 2022.
\newblock Dataset distillation by matching training trajectories.
\newblock In \emph{CVPR}, 4750--4759.

\bibitem[{Chaudhry et~al.(2018{\natexlab{a}})Chaudhry, Dokania, Ajanthan, and Torr}]{chaudhry2018riemannian}
Chaudhry, A.; Dokania, P.~K.; Ajanthan, T.; and Torr, P.~H. 2018{\natexlab{a}}.
\newblock Riemannian walk for incremental learning: Understanding forgetting and intransigence.
\newblock In \emph{ECCV}, 532--547.

\bibitem[{Chaudhry et~al.(2018{\natexlab{b}})Chaudhry, Ranzato, Rohrbach, and Elhoseiny}]{chaudhry2018efficient}
Chaudhry, A.; Ranzato, M.; Rohrbach, M.; and Elhoseiny, M. 2018{\natexlab{b}}.
\newblock Efficient Lifelong Learning with A-GEM.
\newblock In \emph{ICLR}.

\bibitem[{Chaudhry et~al.(2019)Chaudhry, Rohrbach, Elhoseiny, Ajanthan, Dokania, Torr, and Ranzato}]{chaudhry2019tiny}
Chaudhry, A.; Rohrbach, M.; Elhoseiny, M.; Ajanthan, T.; Dokania, P.~K.; Torr, P.~H.; and Ranzato, M. 2019.
\newblock On tiny episodic memories in continual learning.
\newblock \emph{arXiv preprint arXiv:1902.10486}.

\bibitem[{Deng et~al.(2009)Deng, Dong, Socher, Li, Li, and Fei-Fei}]{deng2009imagenet}
Deng, J.; Dong, W.; Socher, R.; Li, L.-J.; Li, K.; and Fei-Fei, L. 2009.
\newblock Imagenet: A large-scale hierarchical image database.
\newblock In \emph{2009 IEEE conference on computer vision and pattern recognition}, 248--255. Ieee.

\bibitem[{Du et~al.(2023)Du, Jiang, Tan, Zhou, and Li}]{du2023minimizing}
Du, J.; Jiang, Y.; Tan, V.~Y.; Zhou, J.~T.; and Li, H. 2023.
\newblock Minimizing the accumulated trajectory error to improve dataset distillation.
\newblock In \emph{CVPR}, 3749--3758.

\bibitem[{Fernando et~al.(2017)Fernando, Banarse, Blundell, Zwols, Ha, Rusu, Pritzel, and Wierstra}]{fernando2017pathnet}
Fernando, C.; Banarse, D.; Blundell, C.; Zwols, Y.; Ha, D.; Rusu, A.~A.; Pritzel, A.; and Wierstra, D. 2017.
\newblock Pathnet: Evolution channels gradient descent in super neural networks.
\newblock \emph{arXiv preprint arXiv:1701.08734}.

\bibitem[{Ghunaim et~al.(2023)Ghunaim, Bibi, Alhamoud, Alfarra, Al~Kader~Hammoud, Prabhu, Torr, and Ghanem}]{ghunaim2023real}
Ghunaim, Y.; Bibi, A.; Alhamoud, K.; Alfarra, M.; Al~Kader~Hammoud, H.~A.; Prabhu, A.; Torr, P.~H.; and Ghanem, B. 2023.
\newblock Real-time evaluation in online continual learning: A new hope.
\newblock In \emph{CVPR}, 11888--11897.

\bibitem[{Gu et~al.(2023)Gu, Vahidian, Kungurtsev, Wang, Jiang, You, and Chen}]{gu2023efficient}
Gu, J.; Vahidian, S.; Kungurtsev, V.; Wang, H.; Jiang, W.; You, Y.; and Chen, Y. 2023.
\newblock Efficient Dataset Distillation via Minimax Diffusion.
\newblock \emph{arXiv preprint arXiv:2311.15529}.

\bibitem[{Gu et~al.(2022)Gu, Yang, Wei, and Deng}]{gu2022not}
Gu, Y.; Yang, X.; Wei, K.; and Deng, C. 2022.
\newblock Not Just Selection, but Exploration: Online Class-Incremental Continual Learning via Dual View Consistency.
\newblock In \emph{CVPR}, 7442--7451.

\bibitem[{Guo, Liu, and Zhao(2022)}]{guo2022online}
Guo, Y.; Liu, B.; and Zhao, D. 2022.
\newblock Online continual learning through mutual information maximization.
\newblock In \emph{ICML}, 8109--8126. PMLR.

\bibitem[{Harun et~al.(2023)Harun, Gallardo, Hayes, Kemker, and Kanan}]{harun2023siesta}
Harun, M.~Y.; Gallardo, J.; Hayes, T.~L.; Kemker, R.; and Kanan, C. 2023.
\newblock SIESTA: Efficient Online Continual Learning with Sleep.
\newblock \emph{arXiv preprint arXiv:2303.10725}.

\bibitem[{He et~al.(2016)He, Zhang, Ren, and Sun}]{he2016deep}
He, K.; Zhang, X.; Ren, S.; and Sun, J. 2016.
\newblock Deep residual learning for image recognition.
\newblock In \emph{CVPR}, 770--778.

\bibitem[{Kang et~al.(2022)Kang, Mina, Madjid, Yoon, Hasegawa-Johnson, Hwang, and Yoo}]{kang2022forget}
Kang, H.; Mina, R. J.~L.; Madjid, S. R.~H.; Yoon, J.; Hasegawa-Johnson, M.; Hwang, S.~J.; and Yoo, C.~D. 2022.
\newblock Forget-free continual learning with winning subnetworks.
\newblock In \emph{ICLR}, 10734--10750. PMLR.

\bibitem[{Kim et~al.(2022)Kim, Kim, Oh, Yun, Song, Jeong, Ha, and Song}]{kim2022dataset}
Kim, J.-H.; Kim, J.; Oh, S.~J.; Yun, S.; Song, H.; Jeong, J.; Ha, J.-W.; and Song, H.~O. 2022.
\newblock Dataset Condensation via Efficient Synthetic-Data Parameterization.
\newblock \emph{arXiv preprint arXiv:2205.14959}.

\bibitem[{Kirkpatrick et~al.(2017)Kirkpatrick, Pascanu, Rabinowitz, Veness, Desjardins, Rusu, Milan, Quan, Ramalho, Grabska-Barwinska et~al.}]{kirkpatrick2017overcoming}
Kirkpatrick, J.; Pascanu, R.; Rabinowitz, N.; Veness, J.; Desjardins, G.; Rusu, A.~A.; Milan, K.; Quan, J.; Ramalho, T.; Grabska-Barwinska, A.; et~al. 2017.
\newblock Overcoming catastrophic forgetting in neural networks.
\newblock \emph{Proceedings of the national academy of sciences}, 114(13): 3521--3526.

\bibitem[{Krizhevsky, Hinton et~al.(2009)}]{krizhevsky2009learning}
Krizhevsky, A.; Hinton, G.; et~al. 2009.
\newblock Learning multiple layers of features from tiny images.

\bibitem[{Lee et~al.(2017)Lee, Kim, Jun, Ha, and Zhang}]{lee2017overcoming}
Lee, S.-W.; Kim, J.-H.; Jun, J.; Ha, J.-W.; and Zhang, B.-T. 2017.
\newblock Overcoming catastrophic forgetting by incremental moment matching.
\newblock \emph{NeurIPS}, 30.

\bibitem[{Li and Hoiem(2017)}]{li2017learning}
Li, Z.; and Hoiem, D. 2017.
\newblock Learning without forgetting.
\newblock \emph{T-PAMI}, 40(12): 2935--2947.

\bibitem[{Liu et~al.(2023{\natexlab{a}})Liu, Ye, Yu, and Wang}]{liu2023slimmable}
Liu, S.; Ye, J.; Yu, R.; and Wang, X. 2023{\natexlab{a}}.
\newblock Slimmable dataset condensation.
\newblock In \emph{CVPR}, 3759--3768.

\bibitem[{Liu et~al.(2023{\natexlab{b}})Liu, Gu, Wang, Zhu, Jiang, and You}]{liu2023dream}
Liu, Y.; Gu, J.; Wang, K.; Zhu, Z.; Jiang, W.; and You, Y. 2023{\natexlab{b}}.
\newblock DREAM: Efficient Dataset Distillation by Representative Matching.
\newblock In \emph{ICCV}.

\bibitem[{Loo et~al.(2022)Loo, Hasani, Amini, and Rus}]{looefficient}
Loo, N.; Hasani, R.; Amini, A.; and Rus, D. 2022.
\newblock Efficient Dataset Distillation using Random Feature Approximation.
\newblock In \emph{NeurIPS}.

\bibitem[{Lopez-Paz and Ranzato(2017)}]{lopez2017gradient}
Lopez-Paz, D.; and Ranzato, M. 2017.
\newblock Gradient episodic memory for continual learning.
\newblock \emph{NeurIPS}, 30.

\bibitem[{Lu et~al.(2023)Lu, Chen, Zhang, Gu, Zhang, Zhang, Yang, Xuan, Wang, and You}]{lu2023can}
Lu, Y.; Chen, X.; Zhang, Y.; Gu, J.; Zhang, T.; Zhang, Y.; Yang, X.; Xuan, Q.; Wang, K.; and You, Y. 2023.
\newblock Can pre-trained models assist in dataset distillation?
\newblock \emph{arXiv preprint arXiv:2310.03295}.

\bibitem[{Mai et~al.(2021)Mai, Li, Kim, and Sanner}]{mai2021supervised}
Mai, Z.; Li, R.; Kim, H.; and Sanner, S. 2021.
\newblock Supervised contrastive replay: Revisiting the nearest class mean classifier in online class-incremental continual learning.
\newblock In \emph{CVPR}, 3589--3599.

\bibitem[{McCloskey and Cohen(1989)}]{mccloskey1989catastrophic}
McCloskey, M.; and Cohen, N.~J. 1989.
\newblock Catastrophic interference in connectionist networks: The sequential learning problem.
\newblock In \emph{Psychology of learning and motivation}, volume~24, 109--165. Elsevier.

\bibitem[{Nguyen, Chen, and Lee(2020)}]{nguyen2020dataset}
Nguyen, T.; Chen, Z.; and Lee, J. 2020.
\newblock Dataset Meta-Learning from Kernel Ridge-Regression.
\newblock In \emph{ICLR}.

\bibitem[{Nguyen et~al.(2021)Nguyen, Novak, Xiao, and Lee}]{nguyen2021dataset}
Nguyen, T.; Novak, R.; Xiao, L.; and Lee, J. 2021.
\newblock Dataset distillation with infinitely wide convolutional networks.
\newblock \emph{NeurIPS}, 34: 5186--5198.

\bibitem[{Pham, Liu, and Steven(2022)}]{phamcontinual}
Pham, Q.; Liu, C.; and Steven, H. 2022.
\newblock Continual Normalization: Rethinking Batch Normalization for Online Continual Learning.
\newblock In \emph{ICLR}.

\bibitem[{Prabhu et~al.(2023)Prabhu, Cai, Dokania, Torr, Koltun, and Sener}]{prabhu2023online}
Prabhu, A.; Cai, Z.; Dokania, P.; Torr, P.; Koltun, V.; and Sener, O. 2023.
\newblock Online continual learning without the storage constraint.
\newblock \emph{arXiv preprint arXiv:2305.09253}.

\bibitem[{Rebuffi et~al.(2017)Rebuffi, Kolesnikov, Sperl, and Lampert}]{rebuffi2017icarl}
Rebuffi, S.-A.; Kolesnikov, A.; Sperl, G.; and Lampert, C.~H. 2017.
\newblock icarl: Incremental classifier and representation learning.
\newblock In \emph{CVPR}, 2001--2010.

\bibitem[{Ring(1998)}]{ring1998child}
Ring, M.~B. 1998.
\newblock CHILD: A first step towards continual learning.
\newblock In \emph{Learning to learn}, 261--292. Springer.

\bibitem[{Sangermano et~al.(2022)Sangermano, Carta, Cossu, and Bacciu}]{sangermano2022sample}
Sangermano, M.; Carta, A.; Cossu, A.; and Bacciu, D. 2022.
\newblock Sample condensation in online continual learning.
\newblock In \emph{IJCNN}, 01--08. IEEE.

\bibitem[{Shim et~al.(2021)Shim, Mai, Jeong, Sanner, Kim, and Jang}]{shim2021online}
Shim, D.; Mai, Z.; Jeong, J.; Sanner, S.; Kim, H.; and Jang, J. 2021.
\newblock Online class-incremental continual learning with adversarial shapley value.
\newblock In \emph{AAAI}, 9630--9638.

\bibitem[{Shin et~al.(2017)Shin, Lee, Kim, and Kim}]{shin2017continual}
Shin, H.; Lee, J.~K.; Kim, J.; and Kim, J. 2017.
\newblock Continual learning with deep generative replay.
\newblock \emph{NeurIPS}, 30.

\bibitem[{Smith et~al.(2021)Smith, Hsu, Balloch, Shen, Jin, and Kira}]{smith2021always}
Smith, J.; Hsu, Y.-C.; Balloch, J.; Shen, Y.; Jin, H.; and Kira, Z. 2021.
\newblock Always be dreaming: A new approach for data-free class-incremental learning.
\newblock In \emph{ICCV}, 9374--9384.

\bibitem[{Sun et~al.(2022)Sun, Calandriello, Hu, Li, and Titsias}]{suninformation}
Sun, S.; Calandriello, D.; Hu, H.; Li, A.; and Titsias, M. 2022.
\newblock Information-theoretic Online Memory Selection for Continual Learning.
\newblock In \emph{ICLR}.

\bibitem[{Thrun(1998)}]{thrun1998lifelong}
Thrun, S. 1998.
\newblock Lifelong learning algorithms.
\newblock In \emph{Learning to learn}, 181--209. Springer.

\bibitem[{Tiwari et~al.(2022)Tiwari, Killamsetty, Iyer, and Shenoy}]{tiwari2022gcr}
Tiwari, R.; Killamsetty, K.; Iyer, R.; and Shenoy, P. 2022.
\newblock Gcr: Gradient coreset based replay buffer selection for continual learning.
\newblock In \emph{CVPR}, 99--108.

\bibitem[{Vinyals et~al.(2016)Vinyals, Blundell, Lillicrap, Wierstra et~al.}]{vinyals2016matching}
Vinyals, O.; Blundell, C.; Lillicrap, T.; Wierstra, D.; et~al. 2016.
\newblock Matching networks for one shot learning.
\newblock \emph{Advances in neural information processing systems}, 29.

\bibitem[{Wang et~al.(2023)Wang, Gu, Zhou, Zhu, Jiang, and You}]{wang2023dim}
Wang, K.; Gu, J.; Zhou, D.; Zhu, Z.; Jiang, W.; and You, Y. 2023.
\newblock DiM: Distilling Dataset into Generative Model.
\newblock \emph{arXiv preprint arXiv:2303.04707}.

\bibitem[{Wang et~al.(2022)Wang, Zhao, Peng, Zhu, Yang, Wang, Huang, Bilen, Wang, and You}]{wang2022cafe}
Wang, K.; Zhao, B.; Peng, X.; Zhu, Z.; Yang, S.; Wang, S.; Huang, G.; Bilen, H.; Wang, X.; and You, Y. 2022.
\newblock Cafe: Learning to condense dataset by aligning features.
\newblock In \emph{CVPR}, 12196--12205.

\bibitem[{Wang et~al.(2018)Wang, Zhu, Torralba, and Efros}]{wang2018dataset}
Wang, T.; Zhu, J.-Y.; Torralba, A.; and Efros, A.~A. 2018.
\newblock Dataset distillation.
\newblock \emph{arXiv preprint arXiv:1811.10959}.

\bibitem[{Xiang et~al.(2019)Xiang, Fu, Ji, and Huang}]{xiang2019incremental}
Xiang, Y.; Fu, Y.; Ji, P.; and Huang, H. 2019.
\newblock Incremental learning using conditional adversarial networks.
\newblock In \emph{ICCV}, 6619--6628.

\bibitem[{Yoon et~al.(2022)Yoon, Madaan, Yang, and Hwang}]{yoon2022online}
Yoon, J.; Madaan, D.; Yang, E.; and Hwang, S.~J. 2022.
\newblock Online Coreset Selection for Rehearsal-based Continual Learning.
\newblock In \emph{ICLR}.

\bibitem[{Zhang et~al.(2022)Zhang, Pfahringer, Frank, Bifet, Lim, and Jia}]{zhangsimple}
Zhang, Y.; Pfahringer, B.; Frank, E.; Bifet, A.; Lim, N. J.~S.; and Jia, A. 2022.
\newblock A simple but strong baseline for online continual learning: Repeated Augmented Rehearsal.
\newblock In \emph{NeurIPS}.

\bibitem[{Zhao and Bilen(2023)}]{zhao2023dataset}
Zhao, B.; and Bilen, H. 2023.
\newblock Dataset condensation with distribution matching.
\newblock In \emph{WACV}, 6514--6523.

\bibitem[{Zhao, Mopuri, and Bilen(2021)}]{zhao2021dataset}
Zhao, B.; Mopuri, K.~R.; and Bilen, H. 2021.
\newblock Dataset Condensation with Gradient Matching.
\newblock \emph{ICLR}, 1(2): 3.

\end{thebibliography}

\appendix
\newpage

\section{Dataset Details}
More details on the employed online continual learning benchmarks are listed below. 
\begin{itemize}
    \item \textbf{Sequential CIFAR-100} is based on the CIFAR-100 dataset~\cite{krizhevsky2009learning}. 
    It contains 100 classes, each with 500 images for training and 100 images for testing. 
    For each experiment, we randomly separate the dataset into 10 tasks with independent classes. The image size is $32\times 32$.
    \item \textbf{Sequential Mini-ImageNet} is organized from the Mini-ImageNet dataset~\cite{vinyals2016matching}. Mini-ImageNet is originally proposed for meta-learning tasks, where 64, 16 and 20 classes are utilized for training, validating and testing, respectively. We combine all the class and separate each class into 500 images for training and 100 images for testing. The image are resized to $84\times 84$. 
    \item \textbf{Sequential Tiny-ImageNet} splits the Tiny-ImageNet dataset~\cite{deng2009imagenet} into multiple tasks for the setting of continual learning. In this work, we split the original dataset into 20 tasks, each consists of 10 unique tasks. For each class, 500 images are employed for training, and 50 for testing. The image are resized to $64\times 64$. 
\end{itemize}

For each experiment, we randomly sample corresponding number of classes for each task to mitigate the possible influence caused by class order. 
And we report the average results of 10 experiment runs with variance. 

\section{More Implementation Details}
For training the sequential tasks, we adopt the basic training scheme and default settings in~\cite{mai2021supervised}. 
In detail, a reduced ResNet-18~\cite{he2016deep} is adopted as the backbone network, where the output channel number of the first convolutional layer is reduced from 64 to 20. 
The projection head for contrastive learning is a 2-layer Multi-Layer Perceptron (MLP) with the output channel number set as 128. 
An Nearest Class Mean (NCM) classifier proposed in~\cite{mai2021supervised} is adopted for the training of current data and memory data. 
For training, a stochastic gradient decent (SGD) optimizer is adopted with the learning rate set as 0.1. 
At each iteration, a mini-batch of size 10 is obtained from the stream data in the current task, and the mini-batch size for memory replay is 100. 

\begin{figure}[t!]
    \centering
    \includegraphics[width=0.35\textwidth]{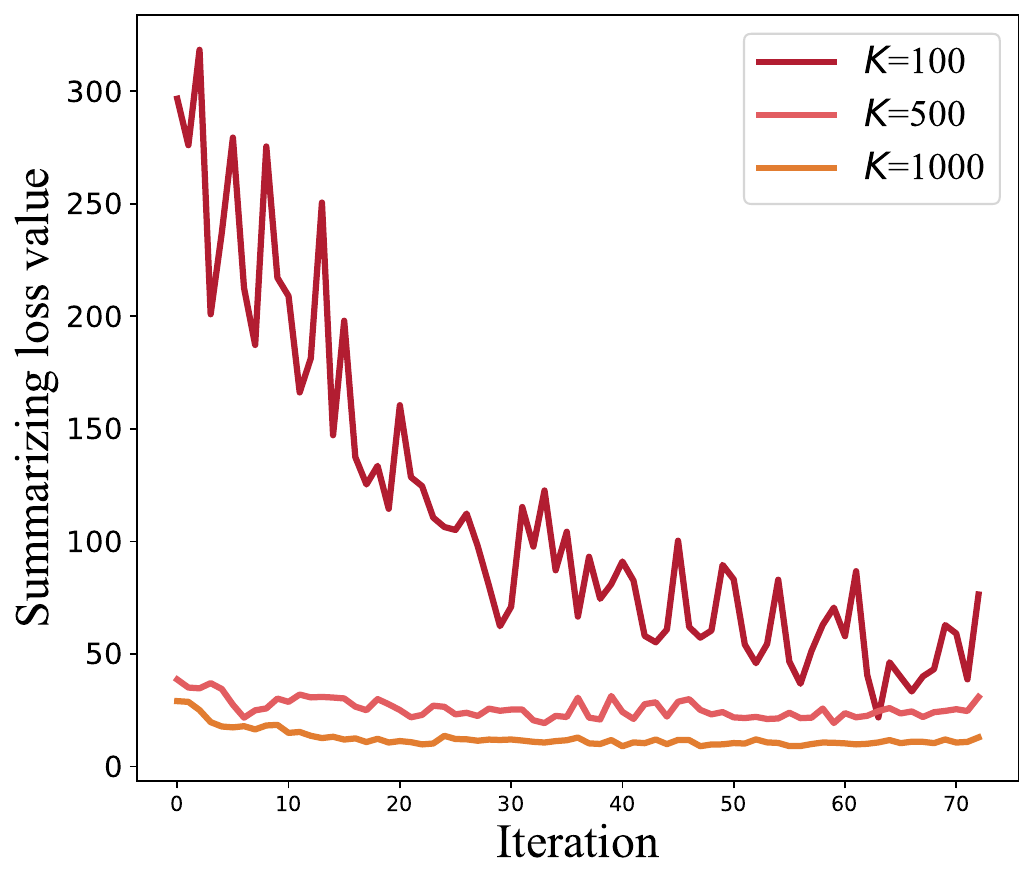}
    \caption{The summarizing loss curve during training under different memory sizes.}
    \label{fig:loss_curve}
\end{figure}

For the optimization of the summarizing images, as the information contained in the 10-sample mini-batch from the stream data is limited, we maintain a queue to store the most recent original samples. 
The buffer size is set as 64 samples per class, which is analyzed in the next section. 
Note that the data in the queue is only utilized for updating the summarized samples, and is unavailable for the sequential-task training. 
The learning rate of the optimizer is set according to the loss scale under different settings. 
We present the loss curve on CIFAR-100 during the training process of different memory sizes in Fig.~\ref{fig:loss_curve}. 
It can be observed that as the summarized sample number of each class increases, the matching loss tends to decrease. 
Therefore, similar to~\cite{kim2022dataset}, we set the learning rate as 2e-4, 1e-3, 4e-3 for 1, 5, 10 images per class, respectively. 

\begin{table}[t!]
    \centering
    \setlength{\tabcolsep}{3pt}
    \scriptsize
    \begin{tabular}{cc|cccccc}
    \toprule
        \multicolumn{2}{c|}{$K$} & SCR & SCR$^{\dagger}$ & RAR & RAR$^{\dagger}$ & RAR-S & RAR-S$^{\dagger}$ \\
         \midrule
        \multirow{2}{*}{2k} & Acc $\uparrow$ & 32.4$_{\pm 0.5}$ & \textbf{34.3$_{\pm 0.5}$} & 30.0$_{\pm 0.6}$ & \textbf{31.5$_{\pm 0.4}$} & 36.5$_{\pm 0.5}$ & \textbf{37.9$_{\pm 0.4}$} \\
        & Fgt $\downarrow$ & 13.6$_{\pm 0.8}$ & \textbf{12.8$_{\pm 0.6}$} & 22.4$_{\pm 0.9}$ & \textbf{21.8$_{\pm 1.0}$} & 24.8$_{\pm 0.5}$ & \textbf{24.1$_{\pm 0.6}$}\\
        \midrule
        \multirow{2}{*}{5k} & Acc $\uparrow$ & 38.3$_{\pm 0.7}$ & \textbf{39.2$_{\pm 0.6}$} & 38.2$_{\pm 0.4}$ & \textbf{39.0$_{\pm 0.5}$} & 45.8$_{\pm 0.5}$ & \textbf{46.6$_{\pm 0.3}$} \\
        & Fgt $\downarrow$ & 7.6$_{\pm 0.5}$ & \textbf{7.3$_{\pm 0.3}$} & 14.6$_{\pm 0.8}$ & \textbf{13.8$_{\pm 0.6}$} & 18.7$_{\pm 0.7}$ & \textbf{17.8$_{\pm 0.4}$}\\
        \bottomrule
    \end{tabular}
    \caption{The performance comparison on larger memory sizes on CIFAR-100 (10 tasks). $^\dagger$ indicates SSD applied. RAR and RAR-S indicates the RAR method applied on ER and SCR, respectively. }
    \label{tab:larger_memory}
\end{table}

\begin{figure}[t!]
    \centering
    \includegraphics[width=0.35\textwidth]{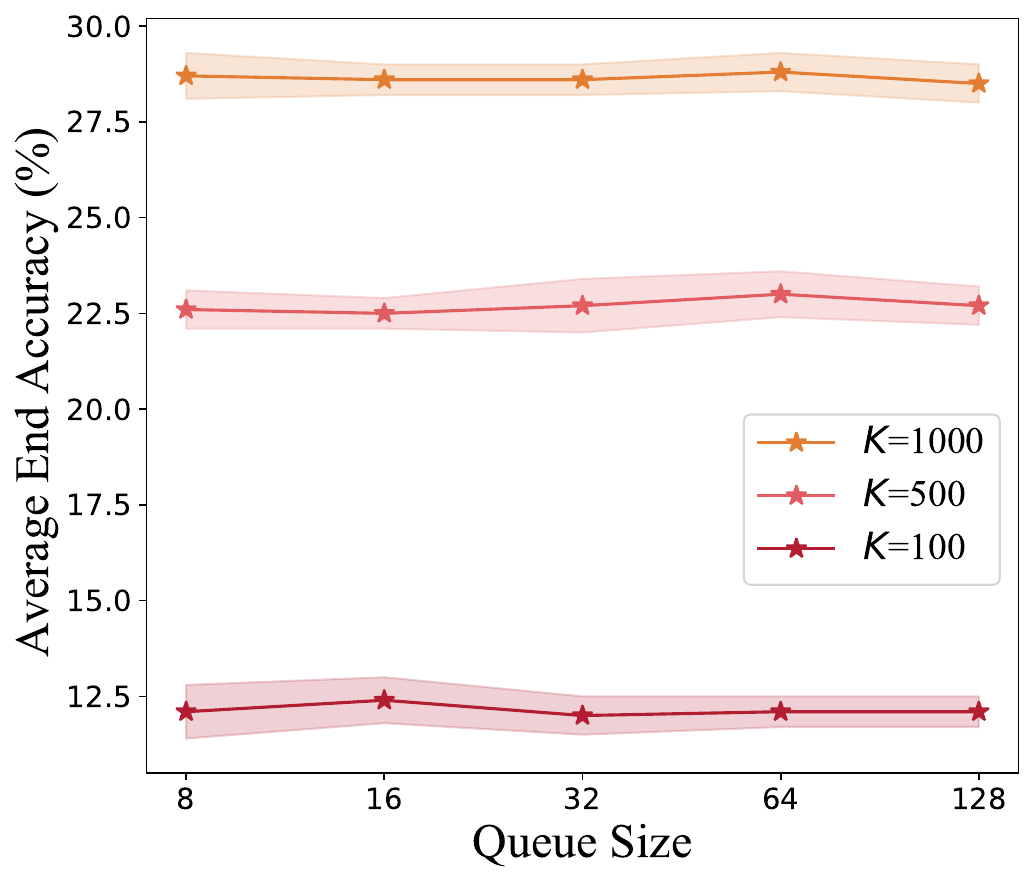}
    \caption{The parameter analysis on the queue size of original images for summarizing.}
    \label{fig:queue}
\end{figure}

\begin{algorithm}[t]
\DontPrintSemicolon
\caption{Summarizing Stream Data }
\label{alg:ssd}
\KwIn{Current task $\mathcal{T}_t$, Memory $\mathcal{M}$, Class set $\mathcal{C}$, Parameters $\theta$, Memory size for a class $k$, summarizing interval $\tau$}
\KwOut{Updated Memory $\mathcal{M}^{'}$}
\SetKwBlock{Begin}{function}{end function}{
Initialize iteration count: $n=0$\;
\For{$\mathcal{B}_t \in \mathcal{T}_t$}{
    Update iteration count: $n=n+1$\;
    \For{$c \in \mathcal{C}$}{
        Acquire $\mathcal{B}_c=\left\{(x_{ti},y_{ti})\in\mathcal{B}_t\vert y_{ti}=c\right\}$\;
        \eIf{
            $\mathcal{M}_c$ is not initialized
        }{
            Randomly select $k$ positions for $\mathcal{M}_c$: $\mathcal{M}_c\leftarrow random(\mathcal{M}_O,k)\leftarrow\mathcal{B}_c$
        }{
            \If{$n\mod\tau=0$}{
                Acquire $\mathcal{M}_c=\{(x_m,y_m)\in\mathcal{M}_S\vert y_m=c\}$\;
                Compute the objective $\mathcal{L}_s$\;
                Update $\mathcal{M}_c$: $\mathcal{M}_c^{*}=\argmin_{\mathcal{M}_c}\mathcal{L}_s$
            }
        }
    }
    Update memory of original images: $\mathcal{M}_O\leftarrow reservior(\mathcal{B}_t)$\;
    Update $\theta$: $\theta\leftarrow\theta-\eta\nabla_\theta\left(\mathcal{L}^{'}_t(\theta;\mathcal{B}_t)+\mathcal{L}^{'}_t(\theta;\mathcal{M}_O)\right)$
}}
\end{algorithm}

\begin{figure}[h!]
\centering
    \begin{subfigure}[t]{0.4\textwidth}
    \centering
    \includegraphics[width=\textwidth]{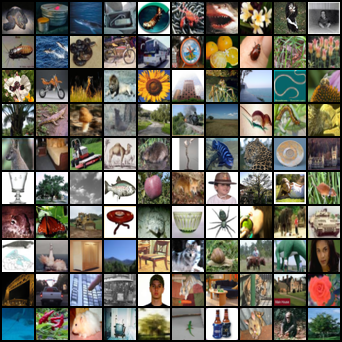}
    \caption{Real images.}
    \end{subfigure}
    \begin{subfigure}[t]{0.4\textwidth}
    \centering
    \includegraphics[width=\textwidth]{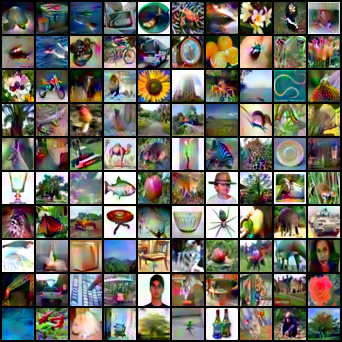}
    \caption{Summarized images.}
    \end{subfigure}
    \caption{The comparison between real images and summarized images. The color, structure and texture features are largely enhanced in the summarized images, indicating richer information. }
    \label{fig:real_sum_comp}
\end{figure}

All the experiments are conducted on a NVIDIA A100 GPU, and the code is implemented in PyTorch version 1.10. 

\section{Pseudo Code}
The detailed pseudo code of the proposed SSD method is presented in Alg.~\ref{alg:ssd}. 

\section{More Ablation Study and Analysis}
\paragraph{Results on larger memory. }
In addition to the experiment results presented in the main manuscript, we apply SSD on SCR and RAR with larger memory sizes and show the results in Tab.~\ref{tab:larger_memory}. 
For better utilization of the auxiliary memory and implementation convenience, the summarizing process does not start until the memory space preserved for the current task is filled. 
Thus, for larger memory sizes, the summarizing iteration is less than that of smaller memory sizes. 
For example, each task in the three benchmarks contains 5,000 images in total. 
Therefore, for the first task, there will be no summarizing process as the training process ends as soon as the original samples fill the memory. 
Under such circumstances, SSD still achieves performance improvement over the high RAR baseline. 
Overall it provides stable improvements on all memory sizes. 

\paragraph{Average forgetting metric.}
We also report the average forgetting metric comparison in Tab.~\ref{tab:larger_memory}. 
As the results are not presented in SCR and RAR, their performance is reproduced by us. 
By comparison, SSD provides less forgetting than the baselines. 

\paragraph{Parameter analysis on the queue size. }
We employ a queue to store the most recent images from the stream data for information summarizing. 
Its influence is analyzed in Fig.~\ref{fig:queue}. 
Under the memory size of 100, the best performance is obtained when queue size is 16. 
For larger memory sizes, the queue size of 64 achieves best performances. 
Generally, the influence on the performance is limited across all queue sizes, and we set 64 as the queue size for all the experiments. 

\section{Visualizations}
We provide more visualizations on the comparison between original images and summarized images in Fig.~\ref{fig:real_sum_comp}. 
The visualization comparison implicitly demonstrate that the color, structure and texture features from the stream data of the corresponding class is integrated into the summarized images. 
Thereby the information contained in the memory is largely improved, and the replay effects are significantly enhanced. 

\section{Limitations and Future Works}
Although summarizing the stream data largely improves the informativeness of the auxiliary memory, there are still some drawbacks of the proposed method. 
The summarizing process involves direct optimization on the images. 
Thus, the optimization space also increases when the memory is able to contain more samples or the image sizes are larger, leading to larger optimization difficulty. 
We aim to explore more reasonable and effective sample summarizing means in the future works.

\end{document}